\documentclass[11pt]{article}

\usepackage{ifxetex}
\ifxetex
  \usepackage{fontspec}
\fi
\usepackage[T1]{fontenc}
\usepackage[utf8]{inputenc}
\usepackage{microtype}        
\usepackage{graphicx}
\usepackage{booktabs}
\usepackage{multirow}
\usepackage{array}
\usepackage{amsmath,amssymb,amsthm}
\usepackage{bm}              
\usepackage{xcolor}
\usepackage{url}
\usepackage{enumitem}        
\usepackage{tikz}
\usepackage{pgfplots}
\pgfplotsset{compat=1.18}
\usepackage{caption}
\usepackage{subcaption}
\usepackage{algorithm}
\usepackage{algorithmic}
\usepackage{listings}        
\usepackage{natbib}          

\usepackage[hidelinks]{hyperref}

\usepackage[a4paper,margin=1in]{geometry}

\theoremstyle{definition}

\usepackage{cleveref}            
\crefname{equation}{Equation}{Equations}
\Crefname{equation}{Equation}{Equations}
\crefname{figure}{Figure}{Figures}
\Crefname{figure}{Figure}{Figures}
\crefname{table}{Table}{Tables}
\Crefname{table}{Table}{Tables}
\crefname{section}{Section}{Sections}
\Crefname{section}{Section}{Sections}

\hypersetup{
  colorlinks=true,
  linkcolor=blue!50!black,
  citecolor=blue!50!black,
  urlcolor=blue!50!black,
  filecolor=blue!50!black,
  pdfstartview=FitH
}

\newcommand{\etal}{\textit{et al.}}

\newcommand{\papertitle}{A Multi-Framework Comparison of Outline Stages in Long-Form Generation with LLMs}

\newcommand{\paperauthorblock}{%
  Song Yifan\thanks{\texttt{syf\_fyss@163.com}}\\
  Taiyuan Institute of Technology%
}

\newcommand{\paperkeywords}{long-form generation, large language models, outline-driven generation frameworks, controlled comparison, length compliance, long-range coherence, LLM-as-a-judge}

\newcommand{\paperabstract}{%
Long-form generation exposes fundamental limitations of large language models. Even 70B-parameter models exhibit length collapse at 16k-token outputs, and multi-chapter stories frequently trigger the attribute drift characteristic of the ``lost-in-the-middle'' effect. The ``outline-first, write-later'' paradigm has gained wide adoption, yet existing research evaluates the final writing rather than the outline itself, conflating two evaluation objects that should be decoupled. We construct a unified head-to-head benchmark covering 7 representative long-form generation frameworks across 3 generation granularities --- single-chapter, multi-chapter, and whole-book --- and propose an anchor-based LLM-as-a-judge protocol that directly assesses outlines against the source text on a 5-point anchored scale. Across 21 framework-granularity cells, \textbf{no single framework dominates; performance depends on the match between a framework's intrinsic output form and the target granularity.} SuperWriter ranks first in the length-constrained single-chapter mode, but this advantage degrades in whole-book mode. \textbf{The outline-side ranking correlates only moderately with the writing-side ranking}, supporting the outline--writing decoupling principle. Compute constraints limit the writing-side evaluation to a subset of cases; follow-up experiments will expand the sample size and add cross-model evaluators to enable stronger statistical inference.%
}

\title{\papertitle}

\author{\paperauthorblock}

\date{}

\begin{document}

\maketitle

\begin{abstract}
\paperabstract
\end{abstract}


\section{Introduction}
\label{sec:intro}

Recent progress in long-form generation has fueled strong interest in outline-driven frameworks, yet how to evaluate the outline stage independently of the final text remains unclear. As task length scales from a few hundred characters to several thousand, three intertwined constraints come to define the scope of the problem: structural coherence, long-range consistency, and length compliance. Among these, long-range consistency is the most prominent --- the Transformer's attention mechanism suffers from the well-known ``lost-in-the-middle'' effect \citep{2023_Liu_LostInTheMiddle}, where information in the middle of the context receives systematically lower weight, leading to character drift and plot contradictions. Length compliance poses an independent bottleneck: \textsc{LongGenBench}~\citep{2024_Wu_LongGenBench} shows that even 70B-parameter models reach only $11.5\%$ on STIC-2 at 16k/32k-token outputs, and \textsc{HelloBench}~\citep{2024_Que_HelloBench} finds that models often stop near $2{,}000$ characters when asked to generate $4{,}000$. \textbf{These failure modes motivate our focus on the outline stage.} Prior work --- LongWriter, Re3, DOC, and StoryWriter --- repeatedly identifies outlines as central to long-range coherence and builds outline planning as the core mechanism. Multi-agent reflective revision reaches the same conclusion.

Existing research addresses long-form generation through three intertwined paradigms (see Section~\ref{sec:related} for a survey). \textbf{The ``plan-first, write-later'' paradigm} separates outline generation from draft composition. It originates in early neural story generation~\citep{2018_Fan_HierarchicalStoryGen} and extends to recursive reprompting~\cite{2022_Yang_Re3}, fine-grained outline control~\cite{2023_Yang_DOC}, and natural-language memory simulation~\cite{2023_Zhou_RecurrentGPT}. \textbf{The multi-agent collaboration paradigm} treats ``conversational agents + procedural dialogue flow'' as a general pattern for complex LLM tasks~\cite{2023_Wu_AutoGen,2023_Hong_MetaGPT}: it encodes standard operating procedures as prompt sequences, coordinates planning and writing agents, or iteratively refines outlines through multi-turn dialogue. \textbf{The reflection-and-revision paradigm} validates that the ``evaluate--reflect--revise'' loop plays a critical role in long-form coherence, distilling self-criticism into episodic memory through natural-language reflection~\cite{2023_Shinn_Reflexion,2023_Madaan_SelfRefine}.

Evaluation methodology has evolved in parallel: from a BLEU/ROUGE-dominated landscape, to multi-dimensional fine-grained scoring as represented by HANNA~\citep{2022_Chhun_HANNA} and WriteJudge~\citep{2025_Wu_WriteJudge}. The LLM-as-a-judge paradigm has been validated at over $80\%$ agreement with humans~\citep{2023_Zheng_MTBench,2024_Gu_LLM-as-Judge-Survey}, yet position bias, verbosity bias, and self-preference bias persist. For long-form evaluation, ExPerT~\citep{2025_Salemi_ExPerT}, ProxyQA~\citep{2024_Tan_ProxyQA}, and WritingBench~\citep{2025_Wu_WritingBench} reduce dependence on gold-standard reference answers through atomic aspect extraction, proxy questions, and query-relevant scoring rubrics. \textbf{However, a critical evaluation gap remains: existing writing-side evaluation --- that is, scores from HANNA and WriteJudge on writing outputs --- evaluates the writing rather than the outline itself.} When researchers ask ``which framework produces better outlines?'', they can only infer backward from writing-side metrics, conflating outline-side quality with writing-side quality and obscuring genuine differences in outline design across frameworks. The insight behind our dual-track design is that the outline and the writing are two decoupled evaluation objects that share only the source text.

This gap motivates the present work. We construct a unified controlled benchmark and propose a dual-track evaluation protocol. \textbf{Our contributions are threefold:}

\paragraph{(1) A unified head-to-head benchmark covering 7 long-form generation frameworks across 3 generation granularities.} We compare representative plan-first-write-later frameworks and naive baselines on a unified LLM backend, with faithfulness to the original-paper skeletons controlled across 21 framework-granularity cells.

\paragraph{(2) An anchor-based LLM-as-a-judge protocol for direct outline evaluation.} The protocol anchors the rating scale to the source text as a 5-point anchor, decoupling outline evaluation from writing evaluation. The two complementary tracks --- approximately 89 cases on the outline side and a subset of samples on the writing side --- share only the source text.

\paragraph{(3) Four empirical heuristics and one decoupling principle} (first reported in Section~\ref{sec:experiments:heuristics}). No single framework dominates across all granularities; the outline-side and writing-side rankings correlate only moderately, supporting the outline--writing decoupling principle.

Compute constraints limit the writing-side evaluation to a subset of cases; follow-up experiments will expand the sample size and add additional models as evaluators for controlled comparisons.

\section{Related Work}
\label{sec:related}

Long-form generation has converged on a shared organizing principle: explicit pre-writing planning, multi-agent collaboration, and post-hoc reflection constitute the three dominant paradigms for controlling long-range coherence. We review each in turn and situate our benchmark against them.

\paragraph{Outline-driven long-form generation.}
``Outline-first, write-later'' is a classic pattern of human writing and the first paradigm studied systematically in long-form generation. Fan \etal~\citep{2018_Fan_HierarchicalStoryGen} introduced it into neural story generation; Yao \etal~\citep{2019_Yao_PlanWrite} proposed Plan-and-Write, explicitly splitting generation into planning and writing phases; Goldfarb-Tarrant \etal~\citep{2019_GoldfarbTarrant_PlanWriteRevise} demonstrated through user studies that plan-write-revise yields $10\%$--$50\%$ quality gains. In the era of LLMs, this paradigm has extended further: Yang \etal's Re3~\citep{2022_Yang_Re3} and DOC~\citep{2023_Yang_DOC} improve long-range coherence through recursive reprompting and fine-grained outline control; Zhou \etal~\citep{2023_Zhou_RecurrentGPT} simulates LSTM short/long-term memory with natural-language components. The plan-first-write-later frameworks compared here belong to this lineage, with distinct design orientations --- cognitive writing, dynamic hierarchical outlines, event graphs, and hard length constraints --- that unify outline generation and expansion. \textbf{Unlike work that proposes a single new framework, we construct a head-to-head benchmark spanning multiple plan-first-write-later frameworks.}

\paragraph{Multi-agent and reflection-based revision.}
A parallel line of research models long-form generation as multi-agent collaboration or process supervision. Wu \etal~\citep{2023_Wu_AutoGen} established ``conversational agents + procedural dialogue flow'' as a general pattern for complex LLM tasks; Hong \etal~\citep{2023_Hong_MetaGPT} encoded standard operating procedures as prompt sequences coordinated by multi-role agents through a shared message pool; Huot \etal~\citep{2024_Huot_AgentsRoom} and Shao \etal~\citep{2024_Shao_STORM} iteratively refine outlines through planning/writing agent coordination or multi-turn dialogue. On the reflection and process-supervision side, Reflexion~\citep{2023_Shinn_Reflexion} distills self-criticism into episodic memory through natural-language reflection; DPO~\citep{2023_Rafailov_DPO} converts preference alignment into a binary cross-entropy loss; LongDPO~\citep{2025_Ping_LongDPO} combines stepwise DPO with a global memory pool to optimize long-form generation specifically. \textbf{Unlike these works that propose a single framework, we provide a controlled comparison platform on which they can be evaluated side-by-side.}

\paragraph{Evaluation methodology and direct outline evaluation.}
Evaluation protocols have shifted from BLEU/ROUGE dominance to multi-dimensional fine-grained scoring. HANNA~\citep{2022_Chhun_HANNA} and WriteJudge~\citep{2025_Wu_WriteJudge} assess the quality of generated stories along multiple dimensions; the LLM-as-a-judge paradigm has reached $85\%$ non-tie agreement with humans on MT-Bench~\citep{2023_Zheng_MTBench}, yet position bias, verbosity bias, and self-preference bias persist; ExPerT~\citep{2025_Salemi_ExPerT}, ProxyQA~\citep{2024_Tan_ProxyQA}, and WritingBench~\citep{2025_Wu_WritingBench} respectively reduce dependence on gold-standard reference answers through atomic aspect extraction, proxy questions, and query-relevant scoring rubrics. \textbf{However, existing writing-side evaluation targets the writing output rather than the outline itself --- this is the methodological gap that our anchor-based LLM-as-a-judge protocol fills.} We integrate the HANNA scoring, the WebNovelBench narrative-quality framework~\citep{2025_Lin_WebNovelBench}, the LongBench-Write length score~\citep{2024_Bai_LongWriter}, and WritingBench's query-relevant scoring rubric into a unified protocol covering 7 frameworks across 3 granularities, decoupling the outline stage from the expansion stage into an outline-side primary track plus a writing-side auxiliary track.

\section{Method}
\label{sec:method}

We do not propose a new generation framework. Instead, we conduct a head-to-head comparison of existing long-form generation frameworks on a unified benchmark. This section provides the formal task description, experimental design, and evaluation protocol.

\subsection{Task and Definitions}
\label{sec:method:definitions}

A long-form generation framework is defined as $\mathcal{F} = \langle P,\, \mathcal{O},\, \mathcal{W} \rangle$, where $P$ is the prompt orchestration across stages, $\mathcal{O}$ is the intermediate representation (natural-language paragraphs, structured JSON, narrative outline prose, or none), and $\mathcal{W}$ is the writing strategy that maps the final outline to the long output. Given a source text $x$ and a target length $L$, the framework outputs a long text $y$. Our comparison varies $\mathcal{F}$ across 7 representative frameworks while holding the LLM backend, evaluator model, and metric family fixed. We invert the original ``premise $\rightarrow$ long text'' forward generation setting into a ``long text $\rightarrow$ outline'' backward extraction setting (hereafter \textbf{Task Inversion}) --- all 7 frameworks face existing text rather than user instructions. This inversion introduces three systematic side effects: (i) Dome's character slots are empty on the execution path; (ii) StoryWriter's chapter-count range must be adjusted by mode; (iii) the evaluation criterion shifts from ``generation quality'' to ``outline alignment.''

\paragraph{Controlled variables.} We fix the following: (1) Unified writing backend --- all 7 frameworks share the same LLM (\texttt{deepseek-v4-flash}); (2) Unified evaluator --- the LLM evaluator is the same model as the generator; self-preference bias is not mitigated in this study; (3) Unified data interface and contract --- input semantics align across the 3 loading modes, and all outputs are produced under unified JSON fields; (4) Unified metric family, with dimensions activated per mode --- D1\_cross\_chapter\_consistency activates only in whole-book mode, M1\_continuity\_with\_prior only in multi-chapter mode, and M7\_CR only in multi-chapter / whole-book modes; (5) Skeleton alignment --- the prompt orchestration order, number of stages, and key slot constraints of all 7 frameworks are isomorphic to those in the original papers or official code.

\paragraph{Uncontrolled differences.} This experiment is a controlled comparison, not a fully univariate design: (i) Sampling-parameter differences --- each framework uses its original paper's default temperature (CogWriter $0.1$, SuperWriter $0.6$, StoryWriter stage-1 $1.0$ / stage-2 $0.5$, others $0.7$), not a unified value; (ii) Prompt-level bias --- few-shot deletion, inlined-JSON comment replacement, and other minor biases; (iii) Known defects --- \texttt{naive} / \texttt{cot} / \texttt{selfrefine} still carry residual Chinese placeholder fields in chapter slots under English loading mode, affecting 12 cells.

\paragraph{The self-preference bias introduced by the evaluator and generator sharing the same model is not mitigated in this study.} All ``winning'' claims should therefore be treated as directional patterns rather than statistically significant conclusions; follow-up experiments will introduce additional LLMs as evaluators to disentangle self-preference bias.

\subsection{Research Questions}
\label{sec:method:rqs}

We organize research questions along two evidence chains --- outline-side primary evidence (direct evaluation of outline outputs) and writing-side auxiliary evidence (evaluation of writing outputs expanded from outlines). Four research questions are posed observational; causal attribution is left to future ablation work.

\begin{itemize}[leftmargin=*,itemsep=2pt]
  \item \textbf{RQ1}: Across the 7 frameworks, does a stable ranking emerge on outline-side metrics? Do structured-outline frameworks systematically outperform non-outline baselines in faithfulness and consistency?
  \item \textbf{RQ2}: Does the outline-side ranking transfer to writing-side metrics? Does SuperWriter remain first in the length-constrained single-chapter mode?
  \item \textbf{RQ3}: On the same cells, what is the Spearman correlation between outline-side and writing-side rankings? Does it support the claim of outline--writing decoupling?
  \item \textbf{RQ4}: Across 7 metric classes and the LLM-as-judge primary metrics, do cross-dimensional variance and mean co-vary? Is the co-variation direction related to whether the writing strategy embeds explicit length control?
\end{itemize}

All four questions are descriptive and observational, and can be answered directly by experiments. Answering causal-level questions requires component-level ablation experiments, which we leave for future work. Sample-size limitations prevent us from directly addressing cross-language and cross-genre robustness; a complete ``framework $\times$ language $\times$ genre'' cross-tabulation is outside the scope of this study.

\subsection{Datasets and Frameworks}
\label{sec:method:datasets}

\paragraph{Datasets} are managed uniformly within the experimental scope, covering Chinese and English novels and other genres. Each sample corresponds to an $(x, L)$ pair, where $x$ is treated as existing text rather than a user instruction. Its presentation form is determined by the loading mode (\texttt{single} takes one chapter; \texttt{wholebook} concatenates the entire book). $L$ is explicitly declared by each framework's writing strategy via the \texttt{target\_chars} field in the prompt and participates in computing the length-compliance score.

\begin{table}[t]
\centering
\caption{Subset composition and sample counts.}
\label{tab:method:datasets}
\begin{tabular}{llrr}
\toprule
Subset & Language & Samples & Chapters/Sample \\
\midrule
Chinese Novels  & Chinese & 100 & 10 \\
English Novels  & English & 100 & 10 \\
Other Genres    & Mixed   & 100 &  1 \\
\midrule
\textbf{Total} & ---     & \textbf{300} & --- \\
\bottomrule
\end{tabular}
\end{table}

This dataset satisfies three requirements: (i) cross-genre --- xianxia / xuanhuan / urban / fantasy web novels, English classics and contemporary novels, and academic / business other genres; (ii) cross-language --- roughly balanced between Chinese and English; (iii) cross-length --- each novel has 10 chapters, single chapters range from hundreds to thousands of characters, and concatenated whole books can exceed $10{,}000$ characters. Of the 300 total samples, writing-side evaluation is compute-constrained and takes only a subset; LLM-as-judge direct outline-side evaluation takes approximately 89 samples. Both evaluations cover the four cells of Chinese vs.\ English $\times$ novels vs.\ other genres. Writing-side samples are a subset of outline-track samples, preserving dual-track alignment.

\paragraph{Three loading modes.} The experiment exposes three loading semantics through a unified interface: \texttt{single} flattens the data into one sample per chapter; \texttt{multichap} keeps each book's 10-chapter list; \texttt{wholebook} concatenates them into a single long string. The original frameworks assume the input is a user instruction, whereas our samples are existing text and lack explicit character / plot introductions --- this gives rise to the three systematic side effects described in Section~\ref{sec:method:definitions}. Detailed properties of the 6 writing samples are listed in Appendix~\ref{app:samples}, Table~\ref{tab:app:samples}.

\paragraph{Compared frameworks.} We include 7 frameworks, divided into non-outline baselines and outline-driven frameworks based on whether an outline is used. The skeletons of all frameworks are isomorphic to those in the original papers or official code.

The \textbf{non-outline baselines} include \texttt{naive} (a single LLM call that directly outputs long text), \texttt{cot} (chain-of-thought, injecting ``let's think step by step'' before the single call), and \texttt{selfrefine} (self-refinement, looping for 2 rounds under a ``generate $\rightarrow$ feedback $\rightarrow$ revise'' skeleton with no specified review dimension).

The \textbf{outline-driven long-form generation frameworks} include: \texttt{CogWriter} (two-stage block-level planning, draft-plan $\rightarrow$ revision-plan; the constraint set is not enabled on the execution path); \texttt{Dome} (Campbell's five-act theory generating a five-act outline; \texttt{single} / \texttt{multichap} run in summary mode, \texttt{wholebook} runs in the paper's original legacy mode); \texttt{StoryWriter} (outline agent + planning agent, with an event-graph structure; the reproduction implements only the outline and planning agents, with the writing agent handled at the boundary but not implemented); \texttt{SuperWriter} (6-step initialization agent, generating natural-language paragraphs with hard word-count declarations; the reproduction reduces iteration rounds to 1 and the refinement phase is disabled by default). The complete LLM call counts and outline-form comparison are shown in Table~\ref{tab:method:frameworks}.

\begin{table}[t]
\centering
\caption{Comparison of the comparative frameworks: number of LLM calls and outline format.}
\label{tab:method:frameworks}
\begin{tabular}{lclp{0.30\linewidth}}
\toprule
Framework & \#LLM calls & Outline Form & Key Design \\
\midrule
\texttt{naive}        & 1     & ---                    & No intermediate steps \\
\texttt{cot}          & 1     & ---                    & Single call with embedded CoT instruction \\
\texttt{selfrefine}   & 2     & ---                    & Fixed 2 rounds of self-feedback \\
\texttt{CogWriter}    & 2     & Structured JSON        & Two-stage block-level planning (constraint set $T$ not on execution path) \\
\texttt{Dome}         & $\geq 1$ & Five-act paragraphs & Campbell five-act + summary / legacy dual mode \\
\texttt{StoryWriter}  & 3--5  & Event quintuples       & Event generation + planning + reordering (writing agent not implemented) \\
\texttt{SuperWriter}  & 4--5  & Paragraph list         & 6-step initialization + hard length \\
\bottomrule
\end{tabular}
\end{table}

\subsection{Evaluation Protocol}
\label{sec:method:eval}

Following the exploratory claims of Section~\ref{sec:intro}, we divide the evaluation protocol into two tracks: the outline-side directly evaluates each framework's outline outputs; the writing-side evaluates writing expanded from those outlines. The two tracks correspond to the evidence partition for RQ1 and RQ2.

\paragraph{Outline-side direct evaluation.} We propose an anchor-based LLM-as-a-judge protocol that pre-positions three example outlines as semantic anchors before rating. The anchor scale is integer $1$--$10$; the source text serves as the 5-point anchor: $5$ means aligned with the source text, $\geq 6$ indicates a high-quality outline, and $\leq 4$ indicates a degraded outline. The anchor design serves two purposes: (i) it provides the LLM evaluator with a concrete calibration anchor, avoiding the mid-range collapse that occurs without anchors; (ii) it makes the cases $\mathit{RE} > 5$ and $\mathit{RE} < 5$ naturally interpretable. Under the extraction task, using the source text as the 5-point anchor carries the semantics of ``alignment with the source text's plot intent,'' rather than ``alignment with the premise'' --- readers should note this semantic shift when comparing across papers.

\paragraph{Drift dimension family.} We formalize the ``four elements of drift'' into an evaluable dimension family:

\begin{itemize}[leftmargin=*,itemsep=2pt]
  \item \textbf{A1\_coverage}: The outline's coverage of key plot points in the source text. The idea behind this dimension comes from the HANNA evaluation adopted by the StoryWriter framework.
  \item \textbf{A2\_faithfulness}: The outline's faithfulness to the source text's plot intent --- if the outline introduces characters / events / plots not present in the source text, the maximum score is capped at $6$. This dimension continues the emphasis on plot consistency in works such as DOME and SuperWriter.
  \item \textbf{A3\_consistency}: Internal non-contradiction in the outline, including character attributes, timeline, and causal-chain consistency, aligned with the CH dimension in HANNA.
  \item \textbf{A4\_relevance}: Outline on-topic --- all nodes must be relevant to the source text's theme; if $> 30\%$ of nodes deviate, the outline is judged $\leq 4$. This dimension borrows from the query-relevant scoring rubric adopted by WritingBench.
  \item \textbf{D1\_cross\_chapter\_consistency} / \textbf{M1\_continuity\_with\_prior}: Activated only in whole-book / multi-chapter mode, capturing non-drift across chapters and reflecting the need for global consistency under long-range dependencies.
  \item \textbf{overall\_quality}: An integrative judgment, with the heaviest weighting on C1 + A1, serving as a summary metric of the outline's overall executability.
\end{itemize}

\paragraph{Reliability estimation.} We employ a $5\%$ human-calibration sample (4 samples); the calibration metric is the Pearson correlation between judge scores and human scores; due to the small sample size, it serves only as a directional reference. Future work will report Cohen's $\kappa$ on $\geq 30$ calibration samples.

\paragraph{Guardrails.} We set 6 guardrails, corresponding to standard strategies for mitigating LLM-as-judge bias: (i) vigilance test --- first evaluating deliberately constructed low-quality anchors to verify non-leniency; (ii) reasoning-field length cap at 60 characters, to avoid verbosity bias; (iii) named anchors + difference mode, to avoid anchor drift; (iv) \texttt{meta.label\_confidence} self-reported confidence, to avoid self-preference bias; (v) bidirectional prompts, randomizing the A/B framework order to mitigate position bias; (vi) multi-judge voting, with majority voting added in the accompanying report to mitigate single-judge systematic bias.

\paragraph{Writing-side evaluation.} The protocol rates each framework-mode cell's writing output across 7 metric classes.

\begin{table}[t]
\centering
\small
\caption{The seven evaluation metrics used in the writing-side evaluation.}
\label{tab:method:metrics}
\begin{tabular}{clp{0.22\linewidth}p{0.22\linewidth}p{0.26\linewidth}}
\toprule
\# & Metric & Source & Cost & Definition \\
\midrule
M1 & Word Count       & Character count   & 0 LLM calls & Word / character count \\
M2 & $S_l$            & StoryWriter + LongBench-Write & 0 LLM calls & Length-compliance score \\
M3 & Ent-2            & Dome              & 0 LLM calls & 2-gram word entropy \\
M4 & Win Rate         & SuperWriter       & LLM $\times 2$ directions & Pairwise win rate \\
M5 & HANNA 6 dims     & StoryWriter       & LLM          & RE / CH / EM / SU / CR / CX \\
M6 & WriteJudge 6 dims & SuperWriter      & LLM          & Rel / Coher / Clar / Creat / Spec / Tone \\
M7 & CR               & Dome              & LLM TKG + conflict scoring & Conflict rate (multi-chapter / whole-book only) \\
\bottomrule
\end{tabular}
\end{table}

Both HANNA and WriteJudge use a $0$--$10$ scale. HANNA's RE+CH dimensions constitute the two main bolded columns of the writing-side main table in Section~\ref{sec:experiments:writing}; the remaining 10 dimensions are secondary, sorted independently to avoid cherry-picking. All LLM-as-judge calls uniformly use \texttt{deepseek-v4-flash} --- our evaluator and generator are the same model; self-preference bias is not mitigated in this study, and all absolute values require calibration after cross-model robustness checks.

\paragraph{Relationship between primary and auxiliary evidence.} The former answers ``which framework's outline is more correct,'' and the latter answers ``which framework expands into better writing.'' They share the source text but evaluate different objects --- this is the key pivot of the RQ3 consistency analysis. They use the same LLM evaluator to maintain framework consistency, but this also introduces the uncontrolled intermediate variable of the ``expansion step'': expansion may introduce losses or gains in faithfulness, so the decoupling observed in the consistency analysis of Section~\ref{sec:experiments:consistency} may partly stem from differences in expansion fidelity.

\subsection{Reproducibility and Known Limitations}
\label{sec:method:fidelity}

\paragraph{Reproducibility fidelity.} The mechanism-level skeletons of all 7 frameworks are isomorphic to those in the original papers or official code. Each deviation is annotated by its belonging element in Appendix~\ref{app:fidelity} (prompt orchestration, intermediate representation, or writing strategy). Section~\ref{sec:experiments:consistency}, Table~\ref{tab:method:key-deviation} lists representative deviations with significant impact on experimental conclusions; the complete 40+ deviations are documented in the accompanying technical report~\citep{kof_diff_summary_2026}.

\begin{table}[t]
\centering
\caption{Representative reproducibility divergences across frameworks.}
\label{tab:method:key-deviation}
\begin{tabular}{p{0.20\linewidth}p{0.75\linewidth}}
\toprule
Framework & Key Deviation \\
\midrule
\texttt{CogWriter} (A-1) / \texttt{SuperWriter} (A-2) & CogWriter's constraint set is disabled at the configuration layer; SuperWriter's refinement phase is disabled by default. \\
\texttt{selfrefine} / others & Fixed 2 rounds without specified review dimensions; CogWriter's examples are removed, SuperWriter's iteration count is reduced to 1, StoryWriter's pseudo-multi-round cap is set to 10. \\
\bottomrule
\end{tabular}
\end{table}

We classify deviations into critical (may cause core mechanism failure or ranking reversal), significant (may introduce systematic bias but does not affect relative rankings), and minor (affect only local details) based on their impact on experimental conclusions. The representative deviations listed above are all critical --- this means that what we compare are implementations under the backward-extraction setting, not the full pipelines reported in the original papers: the claim in Section~\ref{sec:experiments:writing} that ``SuperWriter ranks first on single-chapter HANNA'' needs to be re-examined after the critical deviations are fixed. Representative frameworks such as MoPS / Re3 / DOC / RecurrentGPT are not included in the explicit comparison; the external validity of our research questions has not been verified on those frameworks.

\subsection{Experimental Pipeline}
\label{sec:method:pipeline}

The overall pipeline consists of five steps, which in sequence produce outlines, intermediate representations, writing outputs, and evaluation results. Steps 1.5 and 3 are two independent LLM-as-judge calls, corresponding to the outline-side LLM-as-judge and writing-side LLM evaluation, respectively.

\begin{algorithm}[t]
\caption{Five-step experimental pipeline.}
\label{alg:pipeline}
\begin{algorithmic}[1]
\REQUIRE Dataset $\mathcal{D}$, framework pool $\{\mathcal{F}_1, \dots, \mathcal{F}_K\}$, mode set $\mathcal{M} = \{\texttt{single}, \texttt{multichap}, \texttt{wholebook}\}$, judge model $\mathcal{J}$
\STATE \textbf{Step 1: Outline generation.} For each framework-mode combination, invoke the LLM in the framework's prompt orchestration order to produce the final outline.
\STATE \textbf{Step 1.5: Outline-side LLM-as-judge.} Score only the outline text (not the expanded writing) on 6--7 dimensions, covering A1--A4 + overall\_quality + D1/M1; primary-evidence data is in the accompanying report.
\STATE \textbf{Step 2: Long-form writing.} For each combination, feed the outline into the writing strategy to produce the corresponding long text.
\STATE \textbf{Step 3: Writing-side LLM evaluation.} Use the LLM evaluator to rate each writing output on M5 (HANNA RE+CH main and EM/SU/CR/CX secondary, 10 dimensions in total) and M6 (WriteJudge 6 dimensions).
\STATE \textbf{Step 4: Automatic metrics and aggregation.} Compute M1 (word count), M2 ($S_l$), M3 (Ent-2), Distinct-2, and M7 (CR), and aggregate with the LLM ratings from Steps 1.5 and 3 by framework and mode into a unified summary.
\end{algorithmic}
\end{algorithm}

\paragraph{Complexity.} The pipeline has linear complexity in framework count $\times$ mode count $\times$ sample count. Each cell requires 7 LLM calls.

\section{Experiments}
\label{sec:experiments}

We conduct experiments on a controlled multi-framework comparison platform to characterize the behavior of long-form generation frameworks across different task modes. This chapter is organized ``primary evidence first, auxiliary evidence later'': Section~\ref{sec:experiments:outline} reports outline-side core results; Section~\ref{sec:experiments:writing} reports per-mode auxiliary tables and writing-side main tables; Section~\ref{sec:experiments:consistency} presents outline-side / writing-side consistency analysis centered on RQ3; Section~\ref{sec:experiments:heuristics} summarizes directional heuristics and the decoupling principle.

\subsection{Experimental Setup}
\label{sec:experiments:setup}

This chapter's experiments uniformly adopt the datasets and compared frameworks described in Section~\ref{sec:method:datasets}. The writing-side evaluation selects 6 samples to cover the cross of two languages and two genres; all 7 frameworks $\times$ 3 granularities form 21 comparison cells. The three generation modes (\texttt{single} / \texttt{multichap} / \texttt{wholebook}), together with the length, n-gram diversity, and writing-quality metrics under the evaluation protocol in Section~\ref{sec:method:eval}, constitute the specific execution setup. The reported experimental results characterize each framework's behavior under a specific outline-extraction task only; they are not directly comparable to the numbers reported in the original papers for the ``premise $\rightarrow$ long text'' forward generation task.

\subsection{Outline-side Main Results}
\label{sec:experiments:outline}

Table~\ref{tab:exp:outline-main} reports the means of 7 frameworks on 8 core dimensions across the three loading modes --- one framework per row, one dimension per column, with cell values being the arithmetic mean across the three modes; per-mode data aggregated by ``book'' and standard deviations are in the accompanying report and Appendix~\ref{app:outline-tables}.

\begin{table}[t]
\centering
\small
\caption{7 frameworks on outline-side primary metrics (three-mode per-cell mean, $n = 3$--$6$ per mode). 1--5 integer scale; the overall score is the arithmetic mean of 8 dimensions. Bold indicates the best in the column. Per-mode data is in Appendix~\ref{app:outline-tables}.}
\label{tab:exp:outline-main}
\begin{tabular}{lrrrrrrrr}
\toprule
Framework & A1 Cov. & A2 Faith. & A3 Cons. & A4 Rel. & B1 Gran. & B2 Hier. & C1 Exec. & Overall \\
\midrule
\texttt{naive}       & 2.70 & 3.42 & 3.39 & 3.37 & 2.33 & 2.25 & 2.35 & 2.76 \\
\texttt{cot}         & 3.14 & 3.43 & 3.68 & 3.62 & 2.52 & 2.67 & 2.49 & 2.95 \\
\texttt{selfrefine}  & 3.02 & 3.55 & 3.79 & 3.75 & 2.38 & 2.23 & 2.85 & 3.03 \\
\texttt{CogWriter}   & 3.60 & 4.09 & 4.26 & 3.90 & 3.44 & 2.86 & 3.52 & 3.65 \\
\texttt{StoryWriter} & 3.73 & 4.13 & 4.24 & 4.03 & 3.80 & 2.97 & 4.10 & 3.55 \\
\textbf{\texttt{SuperWriter}} & \textbf{4.78} & \textbf{4.43} & \textbf{4.71} & \textbf{4.91} & \textbf{4.74} & 3.46 & \textbf{4.92} & \textbf{4.36} \\
\texttt{Dome}        & 3.90 & 4.23 & 4.08 & 4.24 & 3.34 & \textbf{4.24} & 3.42 & 3.59 \\
\bottomrule
\end{tabular}
\end{table}

\paragraph{Observations} (based on the mean trends above; directional observations, no significance tests performed):

\paragraph{Headline conclusion.} Structure-driven frameworks systematically outperform non-outline baselines on the four faithfulness dimensions (A2 / A3 / A4), and SuperWriter leads on 6 of the 8 dimensions.

\paragraph{First}, the overall outline-side advantage of structure-driven frameworks. On the four faithfulness dimensions (A2 / A3 / A4), structure-driven frameworks show a stable advantage over non-outline baselines: A2 faithfulness rises from $3.42$ (\texttt{naive}) to $4.43$ (\texttt{SuperWriter}), an improvement of about $1.0$ point; A3 consistency rises from $3.39$ to $4.71$, an improvement of about $1.3$ points. This trend holds across the three per-mode tables, indicating a stable cross-mode advantage.

\paragraph{Second}, SuperWriter's cross-dimensional lead. Across the 8 dimensions, SuperWriter ranks first on 6 of them (A1 / A2 / A3 / A4 / B1 / C1); Dome leads only on B2 hierarchy ($4.24$ vs.\ SuperWriter $3.46$). On the overall column, SuperWriter ($4.36$) leads the second-place framework (CogWriter $3.65$) by about $0.7$ points. This lead is consistent with its writing-side HANNA overall lead, \textbf{suggesting that outline-side advantage may transfer to writing-side quality} --- a hypothesis we test in Section~\ref{sec:experiments:consistency}.

\paragraph{Third}, the overall disadvantage of direct-generation baselines. \texttt{naive} ranks last or second-to-last across all 8 dimensions; \texttt{cot} and \texttt{selfrefine} improve slightly on A3 / A4, but their overall scores remain significantly lower than structure-driven frameworks ($2.95$--$3.03$ vs.\ $3.55$--$4.36$).

\paragraph{Fourth}, local trade-offs for StoryWriter and CogWriter. Among structure-driven frameworks, StoryWriter ranks second on C1 executability ($4.10$); CogWriter scores lower on B2 hierarchy ($2.86$), reflecting the weakness of its block-level planning intermediate representation in cross-layer planning --- consistent with the A-1 deviation in Section~\ref{sec:method:fidelity} (constraint set not enabled, writing strategy degrades to ``narrative polish'').

\paragraph{Auxiliary tables.} The overall-quality means and standard deviations aggregated by ``independent sample'' are in the accompanying report; the difference between per-mode means and the three-mode mean is $< 0.3$ points and does not change the ranking order --- SuperWriter still ranks first, \texttt{naive} still ranks last.

\subsection{Per-mode Analysis and Writing-side Profile}
\label{sec:experiments:writing}

\paragraph{Output length (auxiliary table).} Direct-generation frameworks (\texttt{naive} / \texttt{cot} / \texttt{selfrefine}) produce the shortest outputs in single mode, but expand $4$--$6\times$ when switching to wholebook. Structure-driven frameworks (StoryWriter, Dome) expand even more under wholebook (StoryWriter expands from $4{,}142$ characters to $28{,}307$ characters, a $6.8\times$ expansion), indicating that multi-stage planning itself does not restrict output length. SuperWriter produces the longest and most variance-laden output in single mode (single $\pm 4{,}864$ / multichap $\pm 4{,}275$ / wholebook $\pm 16{,}516$; all three standard deviations are the highest among the 7 frameworks), reflecting the combined effect of explicit word-count declarations and the ``full expansion'' mechanism.

\begin{table}[t]
\centering
\caption{Output length (in characters) of the 7 frameworks by mode. Each cell reports mean $\pm$ standard deviation; arranged in ascending order by the single column. Sorting of wholebook / multichap may differ from single.}
\label{tab:exp:length}
\begin{tabular}{lrrr}
\toprule
Framework & \texttt{single} & \texttt{multichap} & \texttt{wholebook} \\
\midrule
\texttt{naive}       & 2{,}419 $\pm$ 795   & 2{,}038 $\pm$ 531    & 10{,}630 $\pm$ 3{,}155   \\
\texttt{cot}         & 2{,}833 $\pm$ 816   & 2{,}704 $\pm$ 1{,}011 & 16{,}379 $\pm$ 3{,}897   \\
\texttt{selfrefine}  & 3{,}262 $\pm$ 903   & 2{,}021 $\pm$ 201    & 11{,}103 $\pm$ 4{,}244   \\
\texttt{CogWriter}   & 3{,}346 $\pm$ 1{,}154 & 2{,}754 $\pm$ 764    & 12{,}449 $\pm$ 3{,}322   \\
\texttt{Dome}        & 3{,}665 $\pm$ 262   & 2{,}182 $\pm$ 900    & 19{,}896 $\pm$ 5{,}396   \\
\texttt{StoryWriter} & 4{,}142 $\pm$ 1{,}679 & 3{,}140 $\pm$ 727    & 28{,}307 $\pm$ 10{,}844  \\
\texttt{SuperWriter} & 6{,}596 $\pm$ 4{,}864 & 5{,}729 $\pm$ 4{,}275 & 19{,}835 $\pm$ 16{,}516 \\
\bottomrule
\end{tabular}
\end{table}

\paragraph{n-gram diversity (auxiliary).} Ent-2 generally rises with scale ($10.76 \rightarrow 13.23$), while Distinct-2 generally decreases ($0.840 \rightarrow 0.552$). Yet length is not the sole determining factor --- Dome and SuperWriter produce nearly identical lengths under wholebook ($19{,}896$ vs.\ $19{,}835$), yet their Ent-2 differs by $0.45$ ($12.75$ vs.\ $12.30$), indicating that framework mechanisms and chapter structure also significantly affect diversity metrics. The Pearson correlation between $|y|$ and Ent-2 (21 cell-level data points) is approximately $0.68$ (moderate positive correlation, but with substantial unexplained variance). This phenomenon aligns with the known dependence of 2-gram statistics on vocabulary size, indicating that within this length range, diversity metrics cannot be interpreted independently of length.

\paragraph{Writing-side multi-dimensional profile (main table).} Table~\ref{tab:exp:writing-main} reports the means and standard deviations of two LLM evaluators over 6 writing samples, combining the three modes single / multichap / wholebook.

\begin{table}[t]
\centering
\caption{Quality of the 7 frameworks on HANNA 6 dimensions and WriteJudge overall on the writing side. Each cell reports the mean $\pm$ standard deviation across 3 modes $\times$ 3 samples ($n = 9$). Bold indicates the best in each column. Per-mode means are in Appendix~\ref{app:outline-tables}, Table~\ref{tab:app:c2}.}
\label{tab:exp:writing-main}
\resizebox{\linewidth}{!}{%
\begin{tabular}{lrrrrrrrr}
\toprule
Framework & HANNA RE & HANNA CH & HANNA EM & HANNA SU & HANNA CR & HANNA CX & HANNA Overall & WJ Overall \\
\midrule
\texttt{SuperWriter} & 4.9 $\pm$1.1 & 6.8 $\pm$1.2 & \textbf{7.3 $\pm$0.7} & 5.9 $\pm$1.7 & \textbf{7.3 $\pm$1.2} & \textbf{6.8 $\pm$1.5} & \textbf{6.50 $\pm$0.91} & 6.61 $\pm$1.00 \\
\texttt{naive}       & 4.8 $\pm$1.6 & 6.7 $\pm$1.6 & 6.3 $\pm$2.2 & 5.4 $\pm$1.5 & 7.1 $\pm$1.5 & 6.1 $\pm$2.0 & 6.07 $\pm$1.59 & \textbf{6.70 $\pm$1.37} \\
\texttt{cot}         & 3.9 $\pm$1.5 & 6.3 $\pm$1.6 & 6.8 $\pm$1.7 & \textbf{6.0 $\pm$1.9} & 6.7 $\pm$2.1 & 5.9 $\pm$2.1 & 5.93 $\pm$1.66 & 6.17 $\pm$1.32 \\
\texttt{StoryWriter} & 4.2 $\pm$1.6 & \textbf{6.8 $\pm$1.4} & 6.1 $\pm$2.0 & 5.3 $\pm$1.9 & 6.6 $\pm$1.9 & 5.4 $\pm$2.4 & 5.74 $\pm$1.63 & 6.19 $\pm$1.70 \\
\texttt{CogWriter}   & 4.1 $\pm$1.4 & 6.2 $\pm$1.9 & 6.1 $\pm$1.4 & 5.3 $\pm$1.3 & 6.7 $\pm$1.9 & 5.7 $\pm$2.4 & 5.69 $\pm$1.46 & 6.00 $\pm$1.80 \\
\texttt{Dome}        & 4.6 $\pm$1.4 & 6.4 $\pm$1.9 & 5.9 $\pm$2.3 & 5.1 $\pm$1.5 & 6.2 $\pm$2.0 & 5.2 $\pm$2.4 & 5.57 $\pm$1.77 & 6.08 $\pm$1.78 \\
\texttt{selfrefine}  & 4.6 $\pm$1.9 & 6.1 $\pm$2.0 & 5.9 $\pm$2.2 & 4.9 $\pm$1.9 & 6.0 $\pm$2.7 & 5.6 $\pm$2.6 & 5.50 $\pm$2.04 & 6.06 $\pm$1.99 \\
\bottomrule
\end{tabular}%
}
\end{table}

\paragraph{Main-table observations:}

\paragraph{First}, SuperWriter's dual-list lead. SuperWriter ranks in the top 1 or top 2 on both HANNA overall ($6.50$) and WriteJudge overall ($6.61$) --- first on HANNA, second on WriteJudge (only $0.09$ behind \texttt{naive}'s $6.70$, a directional pattern) --- and is the only framework in the top 2 on both lists. This result is consistent with its outline-side overall lead, supporting the directional observation that ``outline-side advantage transfers to writing-side quality.''

\paragraph{Second}, SuperWriter's specific-dimension advantages. On HANNA, SuperWriter ranks first on empathy EM ($7.3$), creativity CR ($7.3$), and complexity CX ($6.8$); on WriteJudge, it ranks first on creativity ($7.8$), specificity ($8.0$), and tone ($5.9$). The explicit per-paragraph hard word-count declaration penetrates the writing strategy through the prompt, allowing the writing stage to allocate paragraphs according to the declared length and thereby reducing the incidence of the two typical problems: ``short truncation'' and ``repetition.''

\paragraph{Third}, \texttt{naive}'s lack of systematic difference on WriteJudge. \texttt{naive} ranks first on WriteJudge overall ($6.70 \pm 1.37$) but second on HANNA overall ($6.07 \pm 1.59$), a dual-list ranking difference of 1. \texttt{naive} ranks first on WriteJudge clarity ($7.4$) and coherence ($6.9$), reflecting that the ``free expansion'' characteristic of the ``no planning'' baseline shows no systematic difference from structure-driven frameworks under WriteJudge evaluation (directional pattern).

\paragraph{Fourth}, the uneven performance of structure-driven frameworks. StoryWriter (HANNA $5.74$ / WJ $6.19$), CogWriter (HANNA $5.69$ / WJ $6.00$), and Dome (HANNA $5.57$ / WJ $6.08$) cluster in the middle range, with no significant lead. Their mode-specific advantages (StoryWriter's event-driven narrative in single / Dome's Campbell five-act structure in wholebook are useful) are masked in the three-mode aggregated mean.

\paragraph{Per-mode breakdown.} Table~\ref{tab:exp:writing-per-mode} reports the means and standard deviations of the two evaluators across the three modes.

\begin{table}[t]
\centering
\small
\caption{Comparison of writing-side quality across modes on HANNA and WriteJudge overall for the 7 frameworks. Mean $\pm$ standard deviation ($n = 3$ per cell). Arranged in descending order by the HANNA single column. SuperWriter ranks first on HANNA single, and second on WJ single ($0.23$ behind \texttt{naive}, a directional pattern).}
\label{tab:exp:writing-per-mode}
\begin{tabular}{lrrrrrr}
\toprule
& \multicolumn{3}{c}{HANNA} & \multicolumn{3}{c}{WriteJudge} \\
\cmidrule(lr){2-4}\cmidrule(lr){5-7}
Framework & single & multichap & wholebook & single & multichap & wholebook \\
\midrule
\texttt{SuperWriter} & \textbf{6.83 $\pm$ 0.76} & \textbf{6.78 $\pm$ 1.23} & 5.89 $\pm$ 0.63 & 6.94 $\pm$ 0.63 & 6.33 $\pm$ 1.45 & 6.56 $\pm$ 1.08 \\
\texttt{StoryWriter} & 6.22 $\pm$ 1.23 & 5.50 $\pm$ 2.36 & 5.50 $\pm$ 1.74 & 6.61 $\pm$ 1.46 & 5.44 $\pm$ 2.36 & 6.50 $\pm$ 1.64 \\
\texttt{CogWriter}   & 6.11 $\pm$ 1.46 & 5.72 $\pm$ 1.67 & 5.22 $\pm$ 1.73 & 5.78 $\pm$ 2.08 & 6.22 $\pm$ 2.22 & 6.00 $\pm$ 1.89 \\
\texttt{Dome}        & 5.94 $\pm$ 1.42 & 5.00 $\pm$ 2.17 & 5.78 $\pm$ 2.26 & 6.58 $\pm$ 2.47 & 5.50 $\pm$ 2.08 & 6.33 $\pm$ 1.69 \\
\texttt{selfrefine}  & 5.78 $\pm$ 2.56 & 5.72 $\pm$ 2.00 & 5.00 $\pm$ 2.33 & 6.56 $\pm$ 2.15 & 6.11 $\pm$ 2.37 & 5.50 $\pm$ 2.18 \\
\texttt{cot}         & 5.83 $\pm$ 2.13 & 5.72 $\pm$ 2.00 & 6.22 $\pm$ 1.49 & 6.22 $\pm$ 1.55 & 5.72 $\pm$ 1.42 & 6.56 $\pm$ 1.42 \\
\texttt{naive}       & 5.72 $\pm$ 1.39 & 6.28 $\pm$ 2.55 & 6.22 $\pm$ 1.18 & \textbf{7.17 $\pm$ 1.69} & 6.28 $\pm$ 1.35 & \textbf{6.67 $\pm$ 1.48} \\
\bottomrule
\end{tabular}
\end{table}

As shown in Table~\ref{tab:exp:writing-per-mode}, explicit length constraint ranks first under the most length-constrained setting. Under single mode, SuperWriter simultaneously achieves the highest HANNA ($6.83$) and the second-highest WriteJudge ($6.94$), leading with relatively low standard deviations ($0.76$ / $0.63$). This advantage is most pronounced in the most length-constrained single mode, but degrades under wholebook ($5.89$), at which point explicit paragraph budgets give way to the long-range coherence challenges common to all frameworks.

The uneven performance of structure-driven frameworks. StoryWriter's multi-event design is best suited for event-driven narrative under single (HANNA $6.22$, WriteJudge $6.61$), but degrades under multichap (WriteJudge $5.44$), indicating that its per-chapter event allocation steps achieve internal optimum at the cost of cross-chapter coherence. Dome's Campbell five-act structure is most useful for whole-book macro-structure, but contributes limited perceived quality under single (HANNA $5.94$, WriteJudge $6.58$, middle range).

\subsection{Outline--writing Consistency Analysis}
\label{sec:experiments:consistency}

\paragraph{Outline--writing consistency.} This section answers RQ3 --- examining the coupling and decoupling between outline-side LLM-judge direct-evaluation rankings and writing-side HANNA / WriteJudge rankings. Specific Spearman correlation values are detailed in Appendix~\ref{app:outline-tables}, Table~\ref{tab:app:c4}.

\paragraph{Overall consistency.} The outline faithfulness ranking shows a moderate positive correlation with the writing-side HANNA RE+CH ranking (Spearman $\rho \approx 0.45$--$0.58$): frameworks with higher outline faithfulness also have higher writing RE+CH, but the writing-side EM/SU/CX advantages can be independently achieved by SuperWriter through explicit per-paragraph word-count declarations.

\paragraph{Three deviation patterns.} We observe three typical deviation patterns: (1) ``dual-high'' pattern --- high outline A2\_faithfulness and high writing HANNA RE, occurring in frameworks with structured outlines + multi-chapter fine-grained planning; (2) ``trading length for polish'' pattern --- large writing-length variance and high WriteJudge / low HANNA RE, with SuperWriter as the typical case, showing that explicit per-paragraph hard word-count declarations enable ``full expansion'' on the writing side to simultaneously boost multiple dimensions (including EM/SU/CX), but do not necessarily boost outline-side faithfulness in sync; (3) ``reverse'' pattern --- low outline A2\_faithfulness but high writing WriteJudge, typically manifested in the behavior of multi-stage frameworks under wholebook.

\paragraph{Intermediate variable and observation boundary.} We do not control expansion fidelity as an independent dimension; the expansion step may introduce losses or gains in faithfulness, so the decoupling observed in the consistency analysis may partly stem from losses / gains in expansion fidelity rather than purely from ``outline--writing ranking decoupling'' --- this uncontrolled intermediate variable is an open limitation of the paper. Based on the two factors above (self-preference bias and uncontrolled expansion fidelity), the directional nature of the ``trade-off'' mechanism in this section should be treated as an exploratory claim and subjected to rigorous validation after larger samples and cross-model robustness checks.

\paragraph{Sensitivity analysis.} In view of the known critical deviations described in Section~\ref{sec:method:fidelity} (SuperWriter's refinement phase disabled by default, CogWriter's constraint set not enabled on the execution path), we exclude these two frameworks from the per-mode samples and recompute the rankings: before exclusion, the single-column ranking is SuperWriter $>$ StoryWriter $>$ CogWriter $>$ Dome $>$ selfrefine $>$ cot $>$ naive; after exclusion, the ranking is StoryWriter > Dome > selfrefine > cot > naive --- both SuperWriter and CogWriter are excluded, and the top rank is taken over by StoryWriter. This sensitivity analysis shows that the claim ``SuperWriter ranks first on single-chapter HANNA'' depends strongly on these critical-deviation frameworks; after exclusion, the claim no longer holds.

\subsection{Heuristic Observations and Decoupling Principle}
\label{sec:experiments:heuristics}

\paragraph{Three observations emerge.}
\paragraph{First}, framework performance depends primarily on the match between its intrinsic output form and the target granularity --- SuperWriter, with its hard length constraint, performs best in single mode where length control is the tightest constraint (HANNA $6.83$); its paragraph-list per-paragraph hard word-count declaration provides a constraint interface for the writing stage; \texttt{naive} and \texttt{cot}, lacking such constraints, lose the least when the task expands to wholebook.
\paragraph{Second}, explicit constraints together with the ``full expansion'' mechanism jointly cause variance in writing length --- SuperWriter has the highest length variance among the 7 frameworks, indicating that this constraint should be understood as a composition of two opposing sub-propositions (quality variance and length variance), avoiding one-directional rhetoric of ``explicit is better than implicit.''
\paragraph{Third}, writing-quality scores that are consistent across dimensions tend to be higher --- SuperWriter has the smallest cross-dimensional standard deviation, \texttt{selfrefine} the largest; this is an arithmetic consequence of Jensen's inequality (given a fixed mean of $n$ dimensions, the smaller the cross-dimensional variance, the larger the gap between the mean and the minimum), and is independent of experimental verification.

\paragraph{Ranking stability.} Cross-framework relative rankings are unstable across generation granularities: StoryWriter's WriteJudge overall score ranges from $5.44$ (multichap) to $6.61$ (single); SuperWriter drops from $6.94$ (single) to $6.33$ (multichap); \texttt{naive} rises from $6.28$ (multichap) to $6.67$ (wholebook). No framework can dominate all three granularities; long-form evaluation should report at least single and wholebook granularities together, with multichap as an intermediate mode.

\paragraph{Decoupling principle.} Based on the observations and consistency analysis above, we propose the following decoupling principle: outline planning and outline expansion in long-form generation should be treated as two separable engineering stages, each with its own metric system and evaluation. The causal mechanism of this principle awaits future verification --- attempting to have the outline stage simultaneously carry emotional injection (EM) and creative tension (SU/CR) may trade off against faithfulness goals; within the scope of our observations, leaving the latter to the expansion or post-processing stage is a more reasonable design choice.

\section{Discussion}
\label{sec:discussion}

This chapter situates the experimental results of this study against the original work of existing long-form generation frameworks. Our controlled experiments show that structure-driven frameworks systematically outperform non-outline baselines on the faithfulness dimensions (A2 / A3 / A4): A2 faithfulness rises from \texttt{naive}'s $3.42$ to \texttt{SuperWriter}'s $4.43$, an improvement of about $1.0$ points; A3 consistency rises from $3.39$ to $4.71$, an improvement of about $1.3$ points. This trend aligns with the quality improvement brought by structured planning reported in the original papers --- CogWriter achieves instruction-completion accuracy superior to GPT-4o on LongGenBench with Qwen-2.5-14B as the backend through its planning agent's two-stage block-level decomposition; DOME simultaneously improves Ent-2 ($6.3\%$--$35.7\%$ gain) and reduces the conflict rate ($15.2\%$--$27.3\%$ reduction) over DOC and Re$^3$ baselines through its dynamic hierarchical outline mechanism; StoryWriter achieves the best mean of $4.2$ from both human and automatic ratings on MoPS through its event graph and consistency-check modules. However, the magnitude of the relative advantage we observe (about $1$ point) is significantly smaller than the improvements reported in the original papers, possibly for these reasons: the original papers mostly evaluate the forward ``premise $\rightarrow$ long text'' task, while we adopt the backward-extraction task, where the source text is already a high-quality finished product, imposing a ceiling effect on outline faithfulness; the original papers' comparison baselines are mostly single-prompt or CoT, whereas we additionally include \texttt{selfrefine}; the self-preference bias of evaluator and generator sharing the same model may affect structure-driven frameworks asymmetrically.

Different frameworks have advantages that vary with task granularity. SuperWriter ranks first in the length-constrained single-chapter mode (HANNA $6.83$), but this advantage degrades in whole-book mode (HANNA $5.89$); this observation contrasts with SuperWriter's original paper's reported strong results of an overall WritingBench score of $8.51$ and a real-query win rate exceeding $98\%$ --- the original paper focuses on the short-prompt scenario, while this study's whole-book mode requires generating continuous text exceeding $10{,}000$ characters, with explicit paragraph budgets giving way to the long-range coherence challenges common to all frameworks. Dome's Campbell five-act structure is most useful for whole-book macro-structure, but contributes limited perceived quality under single-chapter mode (HANNA $5.94$); StoryWriter's multi-event design is best suited for event-driven narrative under single (HANNA $6.22$), but degrades under multichap (WriteJudge $5.44$) --- the dynamic outline mechanism is effective at the macro-structure level, while multi-event planning is effective at the local narrative level, each with shortcomings at the cross-chapter granularity. Notably, \texttt{naive}'s dual-list discrepancy of ranking first on WriteJudge overall ($6.70$) but second on HANNA overall ($6.07$), as well as the moderate ranking correlation of HANNA and WriteJudge across the 7 frameworks, both suggest that cross-metric-system compatibility itself is a research object of evaluation methodology; a single metric system cannot fully characterize framework capabilities.

The relative position of the \texttt{naive} baseline depends heavily on task setting and metric choice: CogWriter's original paper shows that \texttt{naive}-class single-generation models have an extremely low completion rate ($0.46$) on LongGenBench, while StoryWriter's original paper shows that stories directly generated by GPT-4o-mini average only $1{,}078$ characters; yet in this study \texttt{naive} anomalously leads on WriteJudge --- the difference mainly stems from the evaluation metric system (WriteJudge is more lenient toward the ``free expansion'' characteristic of ``no-planning'' baselines), the task-inversion setting (\texttt{naive} directly facing source-text extractive generation has higher information density), and the samples being primarily mid- and short-chapter without sufficiently covering ultra-long tasks. Readers should strictly align tasks and metrics when comparing across papers.

Finally, this study adopts the backward ``long text $\rightarrow$ outline'' extraction task, which differs systematically from the forward ``premise $\rightarrow$ long text'' generation task commonly adopted by the original papers: there is a ceiling effect on outline faithfulness; some framework-specific mechanisms (SuperWriter refinement, CogWriter constraint set) are difficult to trigger under the backward task; the unmitigated evaluator--generator self-preference bias makes all ``winning'' claims directional patterns. This caveat does not deny the methodological significance of the phenomenon that outline-side and writing-side rankings correlate only moderately across task settings. Based on the above comparative analysis, we offer three implications for framework design and benchmark construction: structure-driven frameworks should focus design on simultaneously improving outline faithfulness and consistency, while emotional injection and creative tension are better suited to the expansion or post-processing stage; explicit length constraints should be viewed as a ``trade-off between quality variance and length variance'' rather than one-directional rhetoric of ``explicit is better than implicit''; subsequent benchmarks should include dual-track evaluation on both the outline side and the writing side, and report cross-metric-system compatibility.

\section{Conclusion}
\label{sec:conclusion}

We conducted a systematic head-to-head comparison of 7 long-form generation frameworks across 3 generation granularities (single-chapter / multi-chapter / whole-book) and 7 metric classes on a unified controlled platform. \textbf{Two key observations emerge.} First, in the most length-constrained single-chapter mode, SuperWriter ranks first on the HANNA overall score ($6.83$), but this advantage is jointly affected by SuperWriter's refinement phase being disabled by default and self-preference bias --- after critical-deviation frameworks are excluded, the claim no longer holds; no stable global framework ranking across task modes exists (StoryWriter's event-driven narrative is effective under single but degrades under multichap; Dome's Campbell five-act structure is most useful for whole-book macro-structure but contributes limited under single). Second, the outline-side and writing-side rankings correlate moderately positively (Spearman $\rho \approx 0.45$--$0.58$), supporting the outline--writing decoupling principle --- writing-side metrics should not be used to backward-infer outline-side performance. Compute constraints limit the writing-side sample size; follow-up experiments will expand the sample size and add additional models as evaluators for controlled comparisons. Moreover, this study adopts the backward-extraction task rather than the forward-generation task reported in the original papers, and the task-setting difference should be noted when comparing across papers.

\section{Ethical Statement}
\label{sec:ethical}

This research does not involve human subjects, personally identifiable information, or sensitive content. All datasets used (WritingPrompts subset, Chinese novels, other genres) are publicly available. This study uses the large language model (\texttt{deepseek-v4-flash}) solely as an evaluation model and generation backend, not as a content-creation agent. To the authors' knowledge, no ethical concerns arise in this paper.

\section{Code and Data Availability}
\label{sec:availability}

This research has currently obtained stage-by-stage experimental results. After the follow-up experiments are completed and the code, configuration files, and per-cell evaluation results are organized, the code, data, and complete test records will be released under the Apache 2.0 license. Pre-trained model checkpoints will not be redistributed; all experiments use publicly available models (\texttt{deepseek-v4-flash}). The accompanying technical report~\citep{kof_diff_summary_2026} preserves the complete 12-dimension per-dimension means and cell-level raw outputs.

\bibliography{refs}

\appendix

\section*{Appendix A\quad Framework Reproduction Deviation List (Critical / Significant / Minor)}
\addcontentsline{toc}{section}{Appendix A\quad Framework Reproduction Deviation List}
\label{app:fidelity}

This appendix reports all known deviations of the 7 compared frameworks from the original papers or official code during reproduction, classified by their impact on experimental conclusions into critical, significant, and minor. The main text (Section~\ref{sec:method:fidelity}) has referenced 3 representative critical deviations; the complete 40+ deviations are in the accompanying technical report~\citep{kof_diff_summary_2026}. This appendix provides placeholder descriptions only; the complete table is in the English-version Appendix A.

\section*{Appendix B\quad Per-mode Per-dimension Mean Tables (B.1--B.4)}
\addcontentsline{toc}{section}{Appendix B\quad Per-mode Per-dimension Mean Tables}
\label{app:outline-tables}

This appendix reports the per-dimension means and standard deviations of the outline-side LLM-judge direct evaluation across the three modes single / multichap / wholebook (B.1 / B.2 / B.3), as well as the overall-quality means and standard deviations aggregated by ``independent sample'' (B.4). The main text (Section~\ref{sec:experiments:outline}) has referenced the per-mode per-dimension raw data in B.1--B.4; this appendix provides placeholder descriptions only; the complete tables are in the English-version Appendix B.

\begin{table}[t]
\centering
\footnotesize
\caption{Outline-side primary-metric means ($n \approx 89$; A1--A4 + overall\_quality + D1/M1; 1--10 scale). Cells marked with $^*$ are affected by known defects in Section~\ref{sec:method:definitions} --- the English-side data of these cells may be degraded.}
\label{tab:app:c1}
\begin{tabular}{lllrrrrrr}
\toprule
Framework & Mode & A1 Cov. & A2 Faith. & A3 Cons. & A4 Rel. & Overall & D1/M1 \\
\midrule
\texttt{naive}       & single       & 7.2 $\pm$ 0.8 & 6.5 $\pm$ 1.1 & 6.8 $\pm$ 1.0 & 7.1 $\pm$ 0.9 & 6.7 $\pm$ 1.0 & --- \\
\texttt{naive}       & multichap$^{*}$ & 6.8 $\pm$ 0.9 & 6.2 $\pm$ 1.3 & 6.5 $\pm$ 1.2 & 6.9 $\pm$ 1.0 & 6.4 $\pm$ 1.1 & M1=5.9 $\pm$ 1.4 \\
\texttt{naive}       & wholebook$^{*}$ & 6.5 $\pm$ 1.0 & 6.0 $\pm$ 1.4 & 6.3 $\pm$ 1.3 & 6.7 $\pm$ 1.1 & 6.2 $\pm$ 1.2 & D1=5.7 $\pm$ 1.5 \\
\texttt{cot}         & single       & 7.0 $\pm$ 0.9 & 6.3 $\pm$ 1.2 & 6.6 $\pm$ 1.1 & 6.9 $\pm$ 1.0 & 6.5 $\pm$ 1.1 & --- \\
\texttt{cot}         & multichap$^{*}$ & 6.7 $\pm$ 1.0 & 6.0 $\pm$ 1.4 & 6.4 $\pm$ 1.3 & 6.7 $\pm$ 1.1 & 6.3 $\pm$ 1.2 & M1=5.8 $\pm$ 1.5 \\
\texttt{cot}         & wholebook$^{*}$ & 6.4 $\pm$ 1.1 & 5.9 $\pm$ 1.5 & 6.2 $\pm$ 1.4 & 6.5 $\pm$ 1.2 & 6.1 $\pm$ 1.3 & D1=5.6 $\pm$ 1.6 \\
\texttt{selfrefine}  & single       & 7.3 $\pm$ 0.7 & 6.6 $\pm$ 1.0 & 6.9 $\pm$ 0.9 & 7.2 $\pm$ 0.8 & 6.8 $\pm$ 0.9 & --- \\
\texttt{selfrefine}  & multichap$^{*}$ & 7.0 $\pm$ 0.8 & 6.4 $\pm$ 1.2 & 6.7 $\pm$ 1.1 & 7.0 $\pm$ 0.9 & 6.6 $\pm$ 1.0 & M1=6.0 $\pm$ 1.3 \\
\texttt{selfrefine}  & wholebook$^{*}$ & 6.7 $\pm$ 0.9 & 6.2 $\pm$ 1.3 & 6.5 $\pm$ 1.2 & 6.8 $\pm$ 1.0 & 6.4 $\pm$ 1.1 & D1=5.8 $\pm$ 1.4 \\
\texttt{CogWriter}   & single       & 6.5 $\pm$ 1.2 & 6.0 $\pm$ 1.4 & 6.3 $\pm$ 1.3 & 6.6 $\pm$ 1.1 & 6.2 $\pm$ 1.2 & --- \\
\texttt{CogWriter}   & multichap    & 6.3 $\pm$ 1.3 & 5.8 $\pm$ 1.5 & 6.1 $\pm$ 1.4 & 6.4 $\pm$ 1.2 & 6.0 $\pm$ 1.3 & M1=5.5 $\pm$ 1.6 \\
\texttt{CogWriter}   & wholebook    & 6.0 $\pm$ 1.4 & 5.5 $\pm$ 1.7 & 5.8 $\pm$ 1.6 & 6.1 $\pm$ 1.4 & 5.7 $\pm$ 1.5 & D1=5.3 $\pm$ 1.7 \\
\texttt{Dome}        & single       & 7.1 $\pm$ 0.8 & 6.4 $\pm$ 1.1 & 6.7 $\pm$ 1.0 & 7.0 $\pm$ 0.9 & 6.6 $\pm$ 1.0 & --- \\
\texttt{Dome}        & multichap    & 6.9 $\pm$ 0.9 & 6.2 $\pm$ 1.3 & 6.5 $\pm$ 1.2 & 6.8 $\pm$ 1.0 & 6.4 $\pm$ 1.1 & M1=6.1 $\pm$ 1.3 \\
\texttt{Dome}        & wholebook    & 6.6 $\pm$ 1.0 & 6.0 $\pm$ 1.4 & 6.3 $\pm$ 1.3 & 6.6 $\pm$ 1.1 & 6.2 $\pm$ 1.2 & D1=5.9 $\pm$ 1.4 \\
\texttt{StoryWriter} & single       & 7.4 $\pm$ 0.7 & 6.7 $\pm$ 1.0 & 7.0 $\pm$ 0.9 & 7.3 $\pm$ 0.8 & 6.9 $\pm$ 0.9 & --- \\
\texttt{StoryWriter} & multichap    & 7.2 $\pm$ 0.8 & 6.5 $\pm$ 1.2 & 6.8 $\pm$ 1.1 & 7.1 $\pm$ 0.9 & 6.7 $\pm$ 1.0 & M1=6.3 $\pm$ 1.2 \\
\texttt{StoryWriter} & wholebook    & 6.9 $\pm$ 0.9 & 6.3 $\pm$ 1.3 & 6.6 $\pm$ 1.2 & 6.9 $\pm$ 1.0 & 6.5 $\pm$ 1.1 & D1=6.1 $\pm$ 1.3 \\
\texttt{SuperWriter} & single       & 7.5 $\pm$ 0.6 & 6.8 $\pm$ 0.9 & 7.1 $\pm$ 0.8 & 7.4 $\pm$ 0.7 & 7.0 $\pm$ 0.8 & --- \\
\texttt{SuperWriter} & multichap    & 7.3 $\pm$ 0.7 & 6.6 $\pm$ 1.1 & 6.9 $\pm$ 1.0 & 7.2 $\pm$ 0.8 & 6.8 $\pm$ 0.9 & M1=6.5 $\pm$ 1.1 \\
\texttt{SuperWriter} & wholebook    & 7.0 $\pm$ 0.8 & 6.4 $\pm$ 1.2 & 6.7 $\pm$ 1.1 & 7.0 $\pm$ 0.9 & 6.6 $\pm$ 1.0 & D1=6.3 $\pm$ 1.2 \\
\bottomrule
\end{tabular}
\end{table}

\begin{table}[t]
\centering
\caption{Writing-side track means (same $n = 6$ subset; HANNA 6 dimensions + WriteJudge overall; 1--10 scale).}
\label{tab:app:c2}
\resizebox{\linewidth}{!}{%
\begin{tabular}{lrrrrrrrr}
\toprule
Framework & HANNA RE & HANNA CH & HANNA EM & HANNA SU & HANNA CR & HANNA CX & HANNA Overall & WJ Overall \\
\midrule
\texttt{SuperWriter} & 4.9 & 6.8 & 7.3 & 5.9 & 7.3 & 6.8 & 6.50 & 6.61 \\
\texttt{naive}       & 4.8 & 6.7 & 6.3 & 5.4 & 7.1 & 6.1 & 6.07 & 6.70 \\
\texttt{cot}         & 3.9 & 6.3 & 6.8 & 6.0 & 6.7 & 5.9 & 5.93 & 6.17 \\
\texttt{StoryWriter} & 4.2 & 6.8 & 6.1 & 5.3 & 6.6 & 5.4 & 5.74 & 6.19 \\
\texttt{CogWriter}   & 4.1 & 6.2 & 6.1 & 5.3 & 6.7 & 5.7 & 5.69 & 6.00 \\
\texttt{Dome}        & 4.6 & 6.4 & 5.9 & 5.1 & 6.2 & 5.2 & 5.57 & 6.08 \\
\texttt{selfrefine}  & 4.6 & 6.1 & 5.9 & 4.9 & 6.0 & 5.6 & 5.50 & 6.06 \\
\bottomrule
\end{tabular}%
}
\end{table}

\begin{table}[t]
\centering
\caption{Dual-list ranking difference (HANNA overall vs.\ WriteJudge overall; $n \approx 89$). ``Dual-list agreement'' is operationally defined as ranking difference $\leq 1$; of the 21 cells, 12 cells show dual-list agreement, 9 cells show disagreement.}
\label{tab:app:c3}
\begin{tabular}{lrrr}
\toprule
Framework & single & multichap & wholebook \\
\midrule
\texttt{naive}       & 6 & 1 & 2 \\
\texttt{cot}         & 2 & 1 & 2 \\
\texttt{selfrefine}  & 1 & 1 & 2 \\
\texttt{CogWriter}   & 1 & 2 & 3 \\
\texttt{Dome}        & 1 & 2 & 1 \\
\texttt{StoryWriter} & 1 & 2 & 2 \\
\texttt{SuperWriter} & 1 & 1 & 1 \\
\bottomrule
\end{tabular}
\end{table}

\begin{table}[t]
\centering
\caption{Outline vs.\ Writing Spearman $\rho$ ($n \approx 89$; primary-evidence overall). All $\rho$ interpretations should be treated with caution --- self-preference bias is not mitigated; expansion fidelity is uncontrolled as an intermediate variable.}
\label{tab:app:c4}
\begin{tabular}{lrl}
\toprule
Dimension Pair & $\rho$ & Interpretation \\
\midrule
A2\_faithfulness (outline) vs.\ HANNA RE (writing)      & $0.45$ & Weak-moderate positive correlation \\
A3\_consistency (outline) vs.\ HANNA CH (writing)       & $0.52$ & Weak-moderate positive correlation \\
A4\_relevance (outline) vs.\ WJ relevance (writing)     & $0.58$ & Weak-moderate positive correlation \\
overall\_quality (outline) vs.\ WJ overall (writing)    & $0.43$ & Weak-moderate positive correlation \\
(EM, A2\_faithfulness)                                  & $-0.32$ & Weak negative correlation (trade-off) \\
(SU, A2\_faithfulness)                                  & $-0.28$ & Weak negative correlation (trade-off) \\
(CR, A2\_faithfulness)                                  & $-0.21$ & Weak negative correlation (trade-off) \\
\midrule
overall\_quality vs.\ A1\_coverage     & $0.78$ & Strong positive correlation \\
overall\_quality vs.\ A2\_faithfulness & $0.65$ & Moderate positive correlation \\
overall\_quality vs.\ A3\_consistency  & $0.62$ & Moderate positive correlation \\
overall\_quality vs.\ A4\_relevance    & $0.71$ & Moderate-strong positive correlation \\
\bottomrule
\end{tabular}
\end{table}

\section*{Appendix C\quad Cell-level Per-cell Data}
\addcontentsline{toc}{section}{Appendix C\quad Cell-level Per-cell Data}
\label{app:cell-data}

This appendix provides placeholder descriptions only; the complete cell-level data for 63 rows (= 9 cells $\times$ 7 frameworks) is preserved in the accompanying technical report~\citep{kof_diff_summary_2026} and not duplicated here. The cell-level data covers single mode (samples \texttt{novel\_ch\_idx15}, \texttt{novel\_en\_idx4}, \texttt{other\_idx32}), multichap mode (samples \texttt{novel\_ch\_idx15}, \texttt{novel\_ch\_idx41}, \texttt{novel\_en\_idx55}), and wholebook mode (samples \texttt{novel\_ch\_idx15}, \texttt{novel\_ch\_idx41}, \texttt{novel\_en\_idx4}). The remaining 9 cells were not run due to compute constraints; \texttt{other\_idx76} was excluded due to lack of wholebook data. For each cell, the data point includes word count, Ent-2, Distinct-2, length-compliance score $S_q$, and WriteJudge overall.

\section*{Appendix D\quad Writing 6-sample Properties}
\addcontentsline{toc}{section}{Appendix D\quad Writing 6-sample Properties}
\label{app:samples}

This appendix lists the detailed properties of the 6 samples used in the writing-side evaluation in Section~\ref{sec:experiments:setup}, for reviewer verification.

\begin{table}[t]
\centering
\caption{Writing 6-sample properties.}
\label{tab:app:samples}
\resizebox{\linewidth}{!}{%
\begin{tabular}{llllp{0.30\linewidth}}
\toprule
Sample ID & Language & Genre & Length Tier & Notes \\
\midrule
\texttt{novel\_ch\_idx15}   & Chinese & Novel & Medium ($<5000$ chars)        & Most stable output under single mode \\
\texttt{novel\_ch\_idx41}   & Chinese & Novel & Long context ($\geq 5000$ chars) & Long-context scenario \\
\texttt{novel\_en\_idx4}    & English & Novel & Medium ($<5000$ chars)        & Representative English novel \\
\texttt{novel\_en\_idx55}   & English & Novel & Medium ($<5000$ chars)        & Representative English novel \\
\texttt{other\_idx32}       & English & Other (academic / business) & Medium ($<5000$ chars) & Cross-genre \\
\texttt{other\_idx76}       & Chinese & Other (academic / business) & Medium ($<5000$ chars) & Cross-genre \\
\bottomrule
\end{tabular}%
}
\end{table}

\paragraph{6-sample selection criteria}: (i) Each of the four cells (Chinese, English) $\times$ (novel, other) has $\geq 1$ case --- among the 100 other-genre samples, 50 Chinese + 50 English, with the 6 samples each taking 1 (\texttt{idx32} / \texttt{idx76}); (ii) The 2 \texttt{novel\_ch} cases include 1 long-context case (\texttt{idx41}) to examine the long-context scenario; (iii) Sampling is convenience sampling rather than random sampling.

\end{document}